\documentclass[conference]{IEEEtran}
\IEEEoverridecommandlockouts
\usepackage{cite}
\usepackage{amsmath,amssymb,amsfonts}
\usepackage{algorithmic}
\usepackage{graphicx}
\usepackage{textcomp}
\usepackage{xcolor}
\usepackage{subcaption}
\usepackage{multirow}
\usepackage{booktabs}
\usepackage[colorlinks,
            linkcolor=red,      
            anchorcolor=blue,
            citecolor=green,     
            ]{hyperref}
\usepackage[font=small,labelfont=bf]{caption}
\def\BibTeX{{\rm B\kern-.05em{\sc i\kern-.025em b}\kern-.08em
    T\kern-.1667em\lower.7ex\hbox{E}\kern-.125emX}}
\begin{document}

\title{C-CoT: Counterfactual Chain-of-Thought with Vision-Language Models for Safe Autonomous Driving\\

%\thanks{Identify applicable funding agency here. If none, delete this.}
}

% \author{\IEEEauthorblockN{1\textsuperscript{st} Kefei Tian}
% \IEEEauthorblockA{\textit{College of Transportation} \\
% \textit{Tongji University}\\
% Shanghai, China \\
% 2251090@tongji.edu.cn}
% \and
% \IEEEauthorblockN{2\textsuperscript{nd} Yuansheng Lian}
% \IEEEauthorblockA{\textit{Dept. of Civil Engineering} \\
% \textit{Tsinghua University}\\
% Beijing, China \\
% lys22@mails.tsinghua.edu.cn}
% \and
% \IEEEauthorblockN{3\textsuperscript{rd} Kai Yang}
% \IEEEauthorblockA{\textit{School of Vehicle and Mobility} \\
% \textit{Tsinghua University}\\
% Beijing, China \\
% kaiyang0401@gmail.com}
% \and

% \IEEEauthorblockN{4\textsuperscript{th} Xiangdong Chen}
% \IEEEauthorblockA{\textit{Dept. of Civil and Environmental Engineering} \\
% \textit{National University of Singapore}\\
% Singapore 117576, Singapore \\
% cxd@nus.edu.sg}
% \and
% \IEEEauthorblockN{5\textsuperscript{th} Shen Li}
% \IEEEauthorblockA{\textit{Dept. of Civil Engineering} \\
% \textit{Tsinghua University}\\
% Beijing, China \\
% sli299@tsinghua.edu.cn}
% }

\author{Kefei Tian, Yuansheng Lian, Kai Yang, Xiangdong Chen, and Shen Li$^*$%
\thanks{This work was supported in part by the National Natural Science Foundation of China under Grant 52325209, Grant 52220105, Grant 52272420 and Grant 52502510. \textit{(Corresponding author: Shen Li.)}

Kefei Tian is with the College of Transportation, Tongji University, Shanghai, China (e-mail: 2251090@tongji.edu.cn). Yuansheng Lian and Shen Li are with the Department of Civil Engineering, Tsinghua University, Beijing 100084, China (e-mail: lys22@mails.tsinghua.edu.cn; sli299@tsinghua.edu.cn). Kai Yang is with the School of Vehicle and Mobility, Tsinghua University, Beijing, China (e-mail: kaiyang0401@gmail.com). Xiangdong Chen is with the Department of Civil and Environmental Engineering, National University of Singapore, Singapore 117576 (e-mail: cxd@nus.edu.sg).
}}

\maketitle

\begin{abstract}
Safety-critical planning in complex environments, particularly at urban intersections, remains a fundamental challenge for autonomous driving. Existing methods, whether rule-based or data-driven, frequently struggle to capture complex scene semantics, infer potential risks, and make reliable decisions in rare, high-risk situations.
While vision-language models (VLMs) offer promising approaches for safe decision-making in these environments, most current approaches lack reflective and causal reasoning, thereby limiting their overall robustness. To address this, we propose a counterfactual chain-of-thought (C-CoT) framework that leverages VLMs to decompose driving decisions into five sequential stages: scene description, critical object identification, risk prediction, counterfactual risk reasoning, and final action planning.
Within the counterfactual reasoning stage, we introduce a structured meta-action evaluation tree to explicitly assess the potential consequences of alternative action combinations. This self-reflective reasoning establishes causal links between action choices and safety outcomes, improving robustness in long-tail and out-of-distribution scenarios.
To validate our approach, we construct the DeepAccident-CCoT dataset based on the DeepAccident benchmark and fine-tune a Qwen2.5-VL (7B) model using low-rank adaptation.
Our model achieves a risk prediction recall of 81.9\%, reduces the collision rate to 3.52\%, and lowers L2 error to 1.98 m. Ablation studies further confirm the critical role of counterfactual reasoning and the meta-action evaluation tree in enhancing safety and interpretability.
\end{abstract}


\section{Introduction}

Safety-critical planning in complex urban environments, particularly at dense intersections, remains a fundamental challenge for autonomous driving (AD) \cite{bojarski2016end, teng2023motion}. In these highly interactive scenarios, the presence of occlusions and the uncertain behaviors of surrounding agents demand systems that can understand scene semantics and accurately infer potential traffic risks \cite{tang2022prediction}.

Existing solutions generally fall into two categories: rule-based systems and purely data-driven models. Rule-based methods depend on manually designed heuristics and simplified assumptions, limiting their ability to represent complex scene semantics and agent interactions \cite{aksjonov2021rule, lian2025game}. Purely data-driven approaches, while flexible, often behave as black boxes and require large-scale annotated data, especially for rare high-risk cases \cite{omeiza2021explanations, chen2024end, lian2026bap}. 

vision-language models (VLMs) offer a promising solution to these limitations. By aligning visual observations with high-level language representations, VLMs convert raw perception into structured semantic representations, enabling explicit modeling of scene context and multi-agent relationships \cite{hwang2024emma, sima2024drivelm}. They provide interpretable reasoning processes, improving transparency in safety-critical decision-making \cite{cui2025chain, nie2024reason2drive}. Furthermore, by leveraging large-scale multimodal pre-training, VLMs achieve stronger generalization and robustness in rare and long-tail scenarios \cite{xu2024drivegpt4, shao2024lmdrive, zhou2026opendrivevla}.

Nevertheless, most existing methods remain reactive, generating decisions directly from current observations without explicitly evaluating alternative actions or their consequences \cite{wang2025omnidrive}. The absence of reflective and causal reasoning limits robustness in complex scenarios.
To address this issue, we propose a counterfactual chain-of-thought (C-CoT) framework for risk estimation and safety-critical planning. The key idea is to explicitly guide the model to consider the potential consequences of alternative actions before producing the final decision. 
The main contributions of this paper are summarized as follows:
\begin{itemize}
    \item We propose the C-CoT framework, which incorporates a novel meta-action evaluation tree. This structure enables VLMs to conduct explicit counterfactual reasoning, establishing causal links between high-level action choices and safety outcomes.
    \item We construct the DeepAccident-CCoT dataset based on the DeepAccident benchmark, specifically tailored for VLM-based counterfactual risk estimation and motion planning.
    \item Extensive experiments demonstrate that our approach significantly outperforms existing baselines. Specifically, our C-CoT model, fine-tuned in Qwen2.5-VL (7B), achieves an 81.9\% recall in risk prediction and reduces the collision rate to 3.52\% in complex safety-critical scenarios.
\end{itemize}

\section{Related Work}

\subsection{Risk Reasoning and Safety-Critical Planning for Autonomous Driving}

In autonomous driving, risk reasoning and safety-critical decision-making are fundamental to reliable system deployment. Early approaches are predominantly rule-based, leveraging handcrafted safety metrics such as time-to-collision (TTC), safe distance estimation, and related kinematic models to assess potential hazards, followed by predefined control strategies to mitigate risk \cite{yu2019occlusion, wang2020generating, yang2025interactive}. These methods are interpretable and computationally efficient, but they rely on simplified assumptions and have limited capacity to model complex multi-agent interactions, which restricts their generalization ability.

Data-driven methods learn driving policies directly from data and are typically categorized into imitation learning (IL) and reinforcement learning (RL). IL trains policies by replicating expert demonstrations, enabling human-like behavior when training data adequately covers the target scenarios \cite{acerbo2021safe, acerbo2022mpc}. RL optimizes policies through environment interaction by maximizing cumulative rewards, often incorporating risk-aware costs or constraints to promote safe behavior \cite{hu2026long, lian2026bap, yang2024towards}. While more flexible than rule-based approaches, these methods depend heavily on data quality and coverage, require careful objective design, and often lack interpretability, limiting robustness in rare and safety-critical cases.

Recent work has explored large models to introduce higher-level semantic understanding into risk reasoning \cite{hu2026lift} and safety decision-making \cite{zhou2024vision, long2026vlm}. However, most existing AD systems generate decisions directly from current observations without explicitly evaluating alternative actions and their consequences. This underscores the need for explicit causal and counterfactual reasoning in safety-critical driving scenarios.

\subsection{Visual-Language Models in Autonomous Driving}

In recent years, VLMs \cite{sima2024drivelm, xu2024drivegpt4} have been introduced into autonomous driving by integrating visual encoders with large language models.
Early studies employed VLMs as auxiliary language modules for perception-oriented tasks such as object referring, scene description, and visual question answering (VQA) \cite{xu2024drivegpt4}. In these settings, language outputs primarily enhanced interpretability and were not directly involved in closed-loop control. Subsequent work elevated VLMs to high-level planners that generate semantic driving instructions coordinated with low-level controllers. For instance, VLM-MPC \cite{long2026vlm} and Senna \cite{jiang2024senna} integrate semantic planning with continuous control, improving robustness in complex scenarios.

To further enhance decision reliability, several approaches incorporate chain-of-thought (CoT) reasoning, enforcing structured perception–prediction–planning pipelines \cite{sima2024drivelm}. More recently, counterfactual reasoning has been explored to evaluate alternative actions and their potential consequences, aiming to improve trajectory safety \cite{wang2025omnidrive, peng2025counterfactual}.

However, existing counterfactual approaches remain limited: some generate hypothetical analyses without grounding alternative actions in executable control, while others lack explicit causal modeling from action to risk evolution and final planning outcomes. As a result, semantic reasoning does not consistently translate into reliable decision-making. 
In contrast, our C-CoT framework introduces a structured meta-action evaluation tree to bridge this gap. By explicitly linking hypothetical meta-actions, risk dynamics, and final trajectory generation, our approach enables semantic reasoning processes to directly inform executable decisions, thereby significantly improving robustness.

\section{Methodology}

\subsection{Problem Formulation}

We consider safety-critical decision-making in autonomous driving as a counterfactual risk-aware planning problem. At each time step $t$, the multimodal observation $o_t$ consists of historical information within a past time window of length $T_h$. 
This includes multi-view visual inputs and agent trajectory states, sampled at predefined frequencies. 

Given $o_t$, the objective is to perform C-CoT reasoning and generate safety-aware meta-actions and the corresponding future trajectory.
Specifically, the meta-actions are divided into a short-term decision ($[t, t + T_s]$) and a long-term decision ($[t + T_s, t + T_p]$), where $0 < T_s < T_p$. 

\subsection{Counterfactual Chain-of-Thought Framework}
\label{sec:ccot}

\begin{figure*}[htbp]
    \centering
    \includegraphics[width=0.9\linewidth]{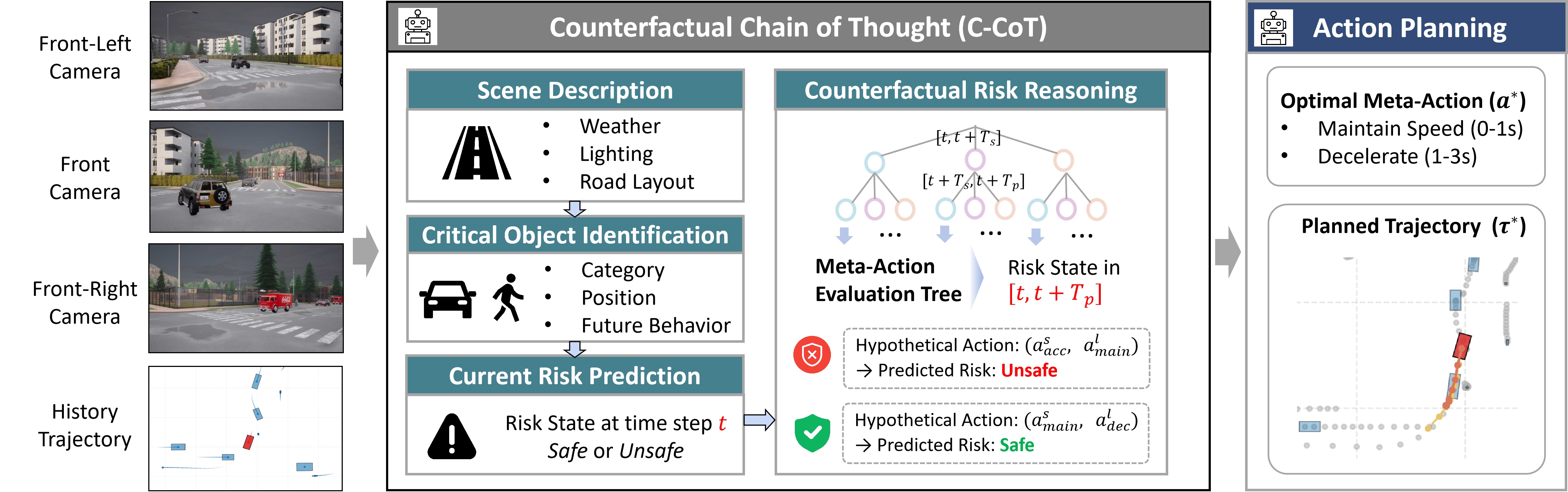}
    \caption{The proposed C-CoT framework. The model takes multi-view camera images and historical trajectories as input, sequentially performing scene description, critical object identification, and current risk estimation. It then evaluates candidate meta-actions through a structured evaluation tree for counterfactual risk reasoning, finally outputting the optimal meta-action and planned trajectory.}
    \label{fig:framework_overview}
\end{figure*}

We model risk perception, counterfactual reasoning, and action planning in autonomous driving as a structured multi-stage reasoning process, formulated as a VQA task.  
Instead of directly mapping observations to actions, the model performs interpretable intermediate reasoning steps, following recent CoT approaches~\cite{wang2025omnidrive, cui2025chain}, as illustrated in Fig.~\ref{fig:framework_overview}:

\begin{enumerate}
    \item \textbf{Scene Description}: Describe the global driving environment, including weather conditions, lighting, and road layout.
    \item \textbf{Critical Object Identification}: Identify potential risk participants by specifying their category, relative position, distance, and possible future behaviors.
    \item \textbf{Current Risk Estimation}: Assess the overall risk level of the scene at the current time step.
    \item \textbf{Counterfactual Risk Reasoning}: Reason about the potential risk evolution over a future horizon of $T_p$ under hypothetical counterfactual actions.
    \item \textbf{Action Planning}: Output both high-level semantic meta-actions and low-level trajectory points. The meta-actions are further divided into short-term actions and long-term actions.
\end{enumerate}

This process links scene understanding, risk dynamics, and motion planning through counterfactual evaluation.
By grounding action selection in such counterfactual analysis, the framework enables reflective safety-aware decision-making and enhances robustness in complex and long-tail scenarios.

\subsection{Counterfactual Meta-Action Evaluation Tree Modeling}

To support counterfactual risk reasoning in the C-CoT framework, we introduce a structured meta-action evaluation tree for simulating counterfactual trajectories and generating ground-truth annotations for counterfactual question answering.

\begin{figure}[t!]
\centering
\includegraphics[width=0.47\textwidth]{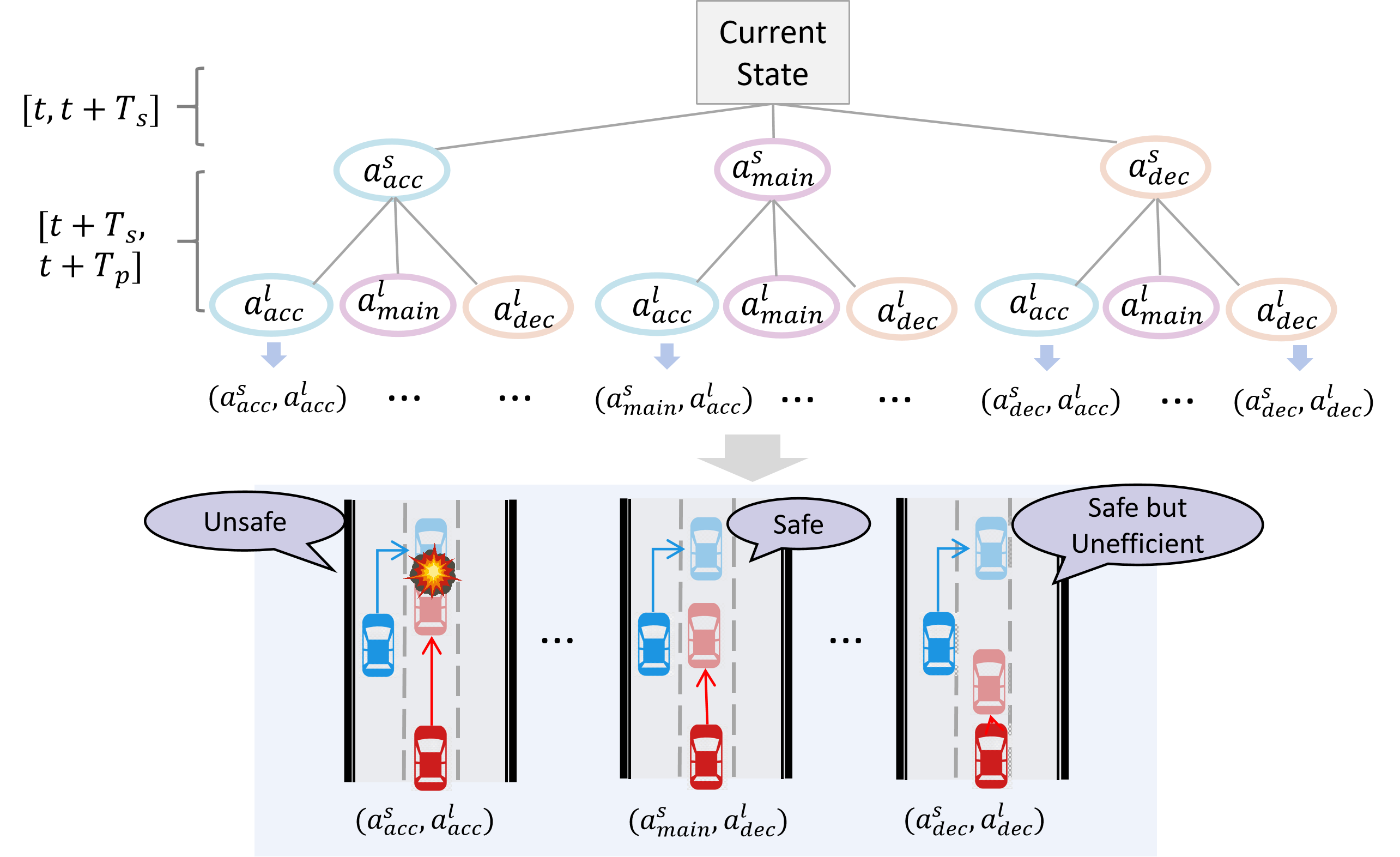} 
\caption{Two-layer meta-action tree illustrating the short-term ($a^s_i$) and long-term ($a^l_i$) decisions. Each root-to-leaf path forms a complete meta-action $(a^s_i, a^l_i)$.}
\label{fig:meta_action_tree}
\end{figure}

\textbf{Meta-Action Tree Construction.}
\label{sec:meta-action tree}
As illustrated in Fig.~\ref{fig:meta_action_tree}, the candidate action space is organized as a two-layer tree. Each first-level node corresponds to a short-term decision $a^s_i$ over $[t, t+T_s]$, and each short-term node branches into long-term decisions $a^l_i$ over $[t+T_s, t+T_p]$. The subscript $i$ indexes discrete behaviors $\{\text{acc}, \text{main}, \text{dec}\}$ for both $a^s_i$ and $a^l_i$. Specifically, \( \text{acc} \) corresponds to acceleration, \( \text{main} \) corresponds to maintaining constant speed, and \( \text{dec} \) corresponds to deceleration. The longitudinal velocity changes are discretized in increments of \( 2 \, \text{m/s}^2 \) \cite{hart2020counterfactual}, and lateral maneuvers remain unchanged. Each root-to-leaf path represents a complete meta-action $(a^s_i, a^l_i)$, and the discrete action space $\mathcal{A}$ is the set of all such paths, forming a compact yet structured evaluation space for counterfactual reasoning.

\textbf{Counterfactual Risk Evaluation.}
Given the observation $o_t$, the model evaluates each candidate action $a = (a^s_i, a^l_i) \in \mathcal{A}$ through counterfactual reasoning: 

\begin{equation}
r(a) = \mathcal{R}(o_t, a; \theta),
\label{eq:reasoning}
\end{equation}
where $\mathcal{R}$ denotes the counterfactual reasoning function parameterized by $\theta$, and $r(a)$ represents the predicted safety outcome associated with action $a$. 
This formulation explicitly captures the causal dependency between staged action choices and future safety states.

\textbf{Tree-Based Action Selection and Planning.}
The optimal action is selected by:

\begin{equation}
a^* = \arg\max_{a \in \mathcal{A}} \mathcal{S}\big(r(a)\big),
\label{eq:selection}
\end{equation}
where $\mathcal{S}$ evaluates the desirability of each counterfactual outcome. 
The selected optimal branch $a^* = (a^{s}_*, a^{l}_*)$ then constrains the subsequent trajectory generation module, producing the final planned trajectory $\tau^*$. 

By systematically traversing the meta-action tree and evaluating the risk evolution, the framework bridges semantic counterfactual analysis and executable motion planning, forming a coherent safety-aware decision pipeline.

\subsection{DeepAccident-CCoT Datasets}
\label{sec:ccot dataset}

We build our training and evaluation sets from the DeepAccident benchmark~\cite{wang2024deepaccident}, which reconstructs safety-critical intersection scenarios from NHTSA pre-crash reports.  
Each sequence contains six surround-view images, environmental metadata, and trajectory annotations of the ego and surrounding vehicles.

\textbf{Temporal Sampling.}
We apply sliding-window sampling over each sequence.  
Each sample contains consecutive frames of $T_h$ seconds, with the last frame as the analysis frame, using a fixed stride to reduce redundancy.  
For normal scenarios, sampling moves forward from stable segments; for collision cases, sampling moves backward from shortly before impact.
After slicing, the dataset contains 2,496 examples, including 948 collision samples.

\textbf{Ground Truth Construction for C-CoT.}
Structured annotations are created for each stage of the five-stage reasoning process:

\begin{enumerate}
    \item \textbf{Scene Description:} Environmental attributes extracted from metadata.
    \item \textbf{Critical Object Identification:} Vehicle with minimum TTC to the ego is selected; in collision cases, the colliding vehicle within the pre-impact window is assigned.
    \item \textbf{Current Risk Estimation:} Frames are labeled \emph{Unsafe} when the TTC is below 3 seconds or when they fall within the collision pre-impact window, while all other frames are labeled \emph{Safe}.
    \item \textbf{Counterfactual Risk Reasoning:} Counterfactual continuous actions are generated through the meta-action evaluation tree (Section~\ref{sec:meta-action tree}), and their safety is evaluated by TTC with surrounding vehicles. The set of nine candidate actions is classified as Safe or Unsafe based on a TTC threshold of 3 seconds.
    \item \textbf{Action Planning:} Ground-truth future ego trajectories supervise planning; for unsafe or collision cases, post-intervention safe trajectories serve as reference.
\end{enumerate}

\textbf{Dataset Split.}
The dataset is split at the scene level (80\% training, 20\% validation) to avoid temporal leakage, preserving similar ratios of collision and non-collision cases.

\section{Experiments}

\subsection{Experimental Setup}

\textbf{Datasets. }
We evaluate our method on the DeepAccident-CCoT dataset described in Section~\ref{sec:ccot dataset}.

\textbf{Implementation Details. }
We build on Qwen-2.5-VL (7B) for risk-aware counterfactual reasoning and trajectory planning. The model processes textual prompts and multimodal inputs: images from three 70$^\circ$ cameras (front, front-left, front-right) at 2\,Hz, and trajectories of ego and nearby vehicles within 30\,m at 10\,Hz, all within a $T_h = 1.5\,\mathrm{s}$ history window.

System, user, and task prompts define the task, scenario, and five-stage C-CoT reasoning. Trajectories are tokenized and combined with visual features in the multimodal Transformer, producing structured JSON outputs for the five-stage reasoning.

The model predicts short-term ($T_s = 1.0\,\mathrm{s}$) and long-term ($T_p = 3.0\,\mathrm{s}$) meta-actions.

\textbf{Training. }  
We fine-tune with supervised low-rank adaptation (LoRA) (rank=8) for 40 epochs using AdamW on 4 A6000 GPUs.

\textbf{Evaluation Metrics. }
We adopt the following metrics for comprehensive evaluation:

\begin{itemize}
    \item \textbf{Language Response Accuracy}: Fraction of correctly completed fields in the structured reasoning outputs.

    \item \textbf{Risk Prediction Accuracy}: Scene-level safe/unsafe classification accuracy and recall.

    \item \textbf{Trajectory L2 Error}: Average Euclidean distance between predicted and ground-truth trajectories at 1\,s and 3\,s.

    \item \textbf{Collision Rate}: Percentage of trajectories that collide within the evaluation horizon.
\end{itemize}

\textbf{Baselines.}
We compare our method with several strong multimodal large language models (LLaVA-1.5 (7B) \cite{liu2024improved}, Llama-3.2-Vision (11B) \cite{lee2025efficient}, 
InternVL-2.5 (8B) \cite{zhu2025internvl3}, DeepSeek-VL (7B) \cite{lu2024deepseekvl}, and Qwen2.5-VL (7B) \cite{yang2025qwen3})
in a zero-shot setting, with prompts adapted to follow the five-step counterfactual reasoning procedure for risk-aware decision-making.

 To evaluate the contribution of each component, we test three variants on the fine-tuned model: 

\begin{itemize}
    \item \textbf{w/o CoT}: Removing the first four C-CoT steps and directly predicting the meta-action and trajectory;

    \item \textbf{w/o Counterfactual-Reasoning}: Skipping counterfactual reasoning and outputting the trajectory after the current risk assessment;

    \item \textbf{w/o Meta-Action}: Removing the meta-action abstraction and directly predicting continuous trajectories.
\end{itemize}

\subsection{Main Results}
Table~\ref{tab:main_results} presents the quantitative comparison between our method and baseline models. Using Qwen2.5-VL as the backbone provides a favorable balance between reasoning capability and model scale. It achieves competitive trajectory performance (L2@3s = 2.38m, Coll. = 5.03\%) with reasonable language accuracy (75.6\%) and risk prediction accuracy (82.3\%), while being more efficient than larger closed-source models such as Llama-3.2-Vision. 

However, as a general-purpose multimodal model, it still shows a noticeable gap in risk prediction and trajectory planning performance. In particular, it tends to underestimate potential hazards, indicating limited sensitivity to subtle safety-critical cues.

With our proposed counterfactual reasoning framework and task-specific fine-tuning, performance improves across all metrics: language accuracy reaches 84.2\%, risk prediction recall rises to 81.9\%, L2@3s decreases to 1.98 m, and the collision rate drops to 3.52\%. The higher risk recall shows that our method can better detect potential hazards, leading to safer trajectory planning and fewer collisions. Overall, these results indicate stronger risk awareness and more reliable safety-critical decision-making.

\begin{table}[t]
\centering
\caption{\textbf{Quantitative comparison on the DeepAccident validation set.}}
\label{tab:main_results}
\scriptsize
\setlength{\tabcolsep}{5pt}
\renewcommand{\arraystretch}{1.0}
\begin{tabular}{lcccccc}
\toprule
& \multicolumn{1}{c}{Language} 
& \multicolumn{2}{c}{Risk Prediction} 
& \multicolumn{3}{c}{Output Trajectory} \\
\cmidrule(lr){2-2}
\cmidrule(lr){3-4}
\cmidrule(lr){5-7}
Method 
& Acc.  
& Acc.  
& Rec. 
& L2@1s 
& L2@3s  
& Coll.  \\
& (\%)$\uparrow$ 
& (\%)$\uparrow$ 
& (\%)$\uparrow$ 
& (m)$\downarrow$ 
& (m)$\downarrow$ 
& (\%)$\downarrow$ \\
\midrule
LLaVA-1.5 (7B)        & 63.5 & 70.2 & 58.1 & 0.76 & 2.92 & 6.84 \\
Llama-3.2-Vision (11B)     & 69.8 & 75.6 & 63.7 & 0.70 & 2.64 & 5.91 \\
InternVL-2.5 (8B)     & 72.4 & 76.9 & 63.5 & 0.63 & 2.35 & 5.01 \\
DeepSeek-VL (7B)      & 74.1 & 78.8 & 69.2 & 0.68 & 2.55 & 5.57 \\
Qwen2.5-VL (7B)       & 75.6 & 82.3 & 68.7 & 0.66 & 2.38 & 5.03 \\
\midrule
\textbf{C-CoT (Ours)}         & \textbf{84.2} & \textbf{85.8} & \textbf{81.9} 
& \textbf{0.54} & \textbf{1.98} & \textbf{3.52} \\
\bottomrule
\end{tabular}
\end{table}

\subsection{Ablation Studies}
\begin{table*}[htbp]
\centering
\caption{\textbf{Ablation study results under different validation settings.}
For positive and mixed samples, lower L2 errors and collision rates indicate better performance. 
For negative samples, since the ground-truth trajectories correspond to collision paths, larger L2 errors reflect safer deviations from the crash trajectory.}
\label{tab:ablation}
\renewcommand{\arraystretch}{0.8}
\begin{tabular}{llcccccc}
\toprule
& & \multicolumn{1}{c}{Language} 
& \multicolumn{2}{c}{Risk Prediction} 
& \multicolumn{3}{c}{Output Trajectory} \\
\cmidrule(lr){3-3}
\cmidrule(lr){4-5}
\cmidrule(lr){6-8}
Validation Type & Model Variant 
& Acc. (\%)$\uparrow$ & Acc. (\%)$\uparrow$ & Rec. (\%)$\uparrow$
& L2@1s (m)$\downarrow$ & L2@3s (m)$\downarrow$ & Coll. (\%)$\downarrow$\\
\midrule
\multirow{4}{*}{Positive Samples}
& w/o CoT & N/A & N/A & N/A & 0.35 & 1.85 & 2.83 \\
& w/o Counterfactual Reasoning & 85.9 & 83.8 & N/A & \textbf{0.32} & 1.61 & 1.31 \\
& w/o Meta-Action Abstraction & 86.3 & 93.4 & N/A & 0.33 & 1.73 & 1.45 \\
& {Full Model}  & \textbf{86.5} & \textbf{93.8} & \textbf{N/A} 
& \textbf{0.32} & \textbf{1.58} & \textbf{1.12} \\
\midrule
\multirow{4}{*}{Negative Samples}
& w/o CoT & N/A & N/A & N/A & 0.78 & 2.87 & 10.39 \\
& w/o Counterfactual Reasoning & \textbf{83.5} & 77.1 & 70.8 & 1.15 & 3.68 & 8.92 \\
& w/o Meta-Action Abstraction & 82.9 & \textbf{79.3} & 81.5 & 0.98 & 2.94 & 9.13 \\
& {Full Model}  & {83.4} & 78.2 & \textbf{81.6} 
& \textbf{1.19} & \textbf{3.72} & \textbf{8.64} \\
\midrule
\multirow{4}{*}{Mixed Samples}
& w/o CoT & N/A & N/A & N/A & 0.58 & 2.35 & 4.72 \\
& w/o Counterfactual Reasoning & 81.8 & 83.5 & 78.0 & 0.55 & 2.03 & 3.69 \\
& w/o Meta-Action Abstraction & 82.5 & 85.0 & 81.5 & 0.56 & 2.14 & 3.95 \\
& {Full Model}  & \textbf{84.2} & \textbf{85.8} & \textbf{81.9} 
& \textbf{0.54} & \textbf{1.98} & \textbf{3.52} \\
\bottomrule
\end{tabular}
\end{table*}

To analyze the contribution of each key component, we conduct ablation experiments by progressively removing CoT, counterfactual reasoning, and meta-action abstraction. The quantitative results are shown in Table~\ref{tab:ablation}.

Removing the \textbf{CoT} leads to consistent performance degradation. 
In mixed samples, $L2@3s$ and collision rate increase ($1.98 \to 2.35\,\text{m}$, $3.52\% \to 4.72\%$); 
in negative samples, collision rate rises ($8.64\% \to 10.39\%$). 
This shows that explicit reasoning enhances planning stability, safety, and interpretability.

Removing the \textbf{Counterfactual Reasoning} reduces risk prediction and safety performance. 
In mixed samples, recall drops ($81.9\% \to 78.0\%$) and collision rate rises ($3.52\% \to 3.69\%$). 
These demonstrate that the counterfactual mechanism effectively improves both accuracy and recall in risk prediction, particularly by enhancing sensitivity to high-risk cases and increasing the proportion of unsafe predictions. 
By simulating what would happen if a certain action were taken, the model proactively avoids potentially hazardous behaviors, thereby improving overall safety.

Removing the \textbf{Meta-Action Abstraction} degrades trajectory planning. 
In positive samples, $L2@3s$ and collision rate increase ($1.58 \to 1.73\,\text{m}$, $1.12\% \to 1.45\%$); 
in negative samples, collision rate rises ($8.64\% \to 9.13\%$). 
The high-level Meta-Action abstraction serves as a structured bridge between risk reasoning and low-level continuous control, enabling more accurate and safer trajectory outputs. 
Overall, all three components contribute substantially to performance improvements.

\subsection{Qualitative Analysis}

\begin{figure}[t]
    \centering
    
    \begin{subfigure}{\linewidth}
        \centering
        \includegraphics[width=\linewidth]{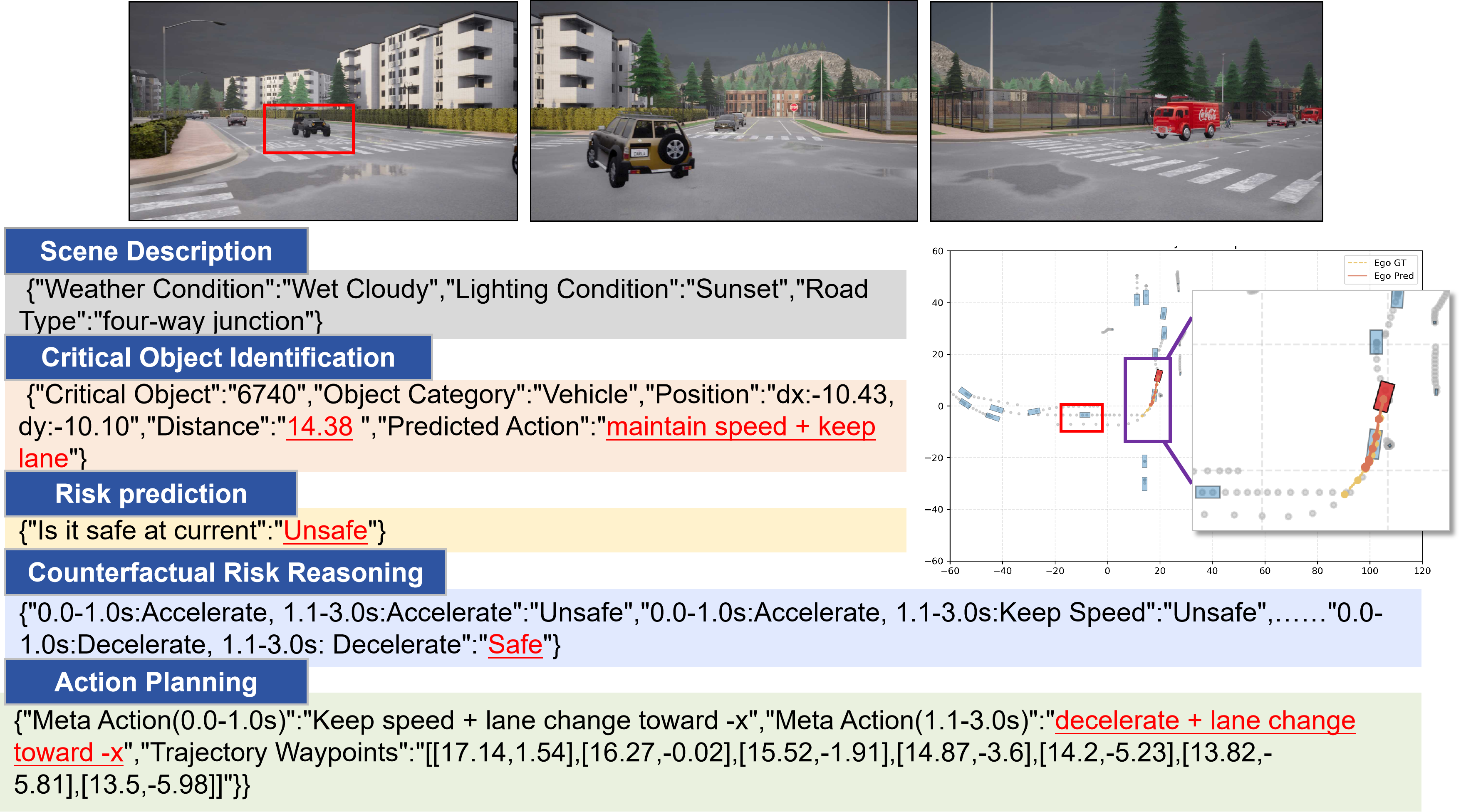}
        \caption{Unsafe driving scenario. The model predicts the scene as \textit{Unsafe} and selects deceleration, generating a safe trajectory that avoids the collision path.}
        \label{fig:qual_collision}
    \end{subfigure}
    
    \vspace{0.5em}
    
    \begin{subfigure}{\linewidth}
        \centering
        \includegraphics[width=\linewidth]{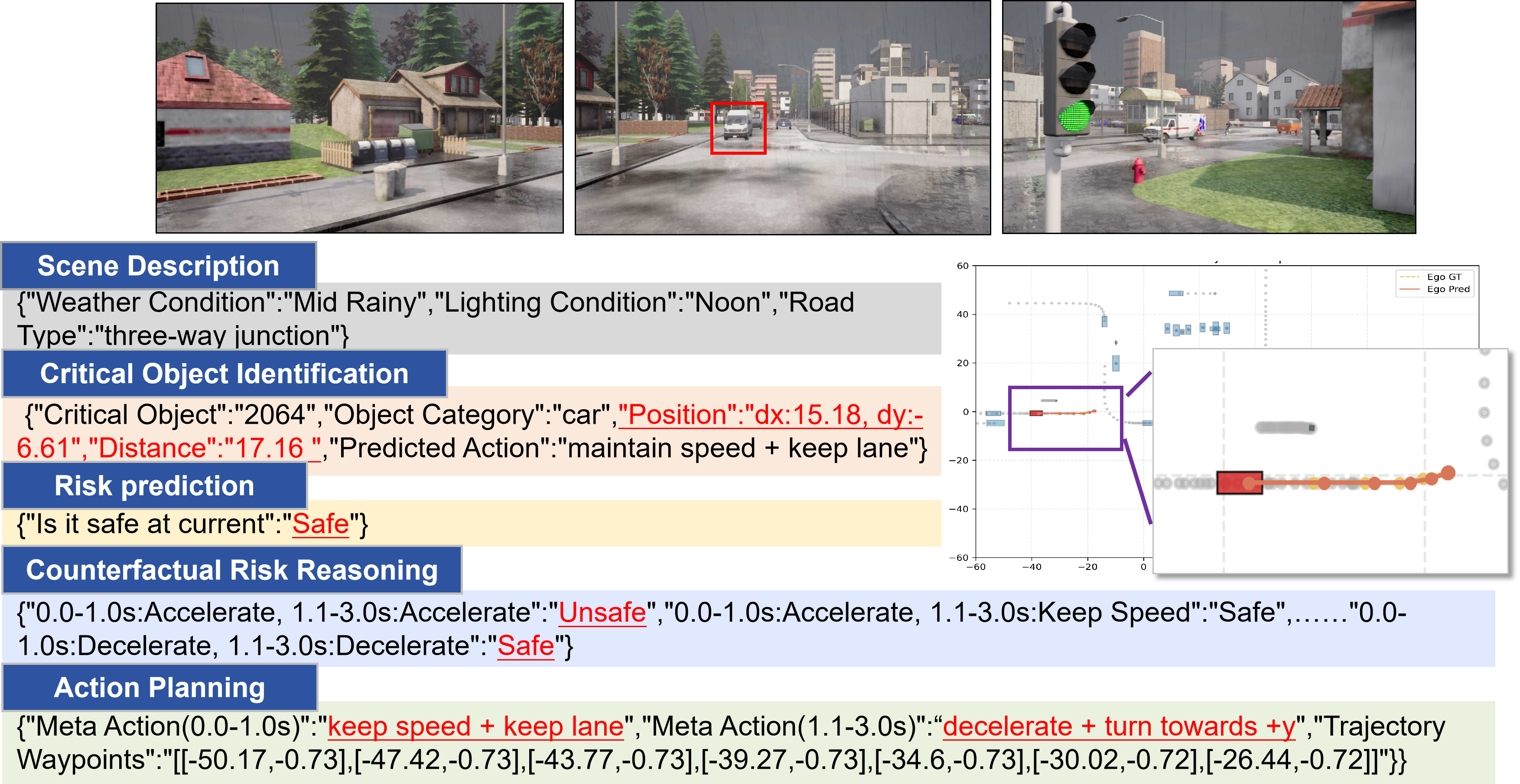}
        \caption{Safe driving scenario. The model balances safety and efficiency via counterfactual reasoning, producing a trajectory close to the ground truth.}
        \label{fig:qual_safe}
    \end{subfigure}
    
    \caption{\textbf{Qualitative results of the proposed C-CoT framework.} 
    (a) In a collision-prone scenario, counterfactual reasoning leads to deceleration and successful collision avoidance. 
    (b) In a safe scenario, the model achieves stable and efficient planning with trajectories closely matching the ground truth.}
    
    \label{fig:qualitative}
\end{figure}

We present two representative cases to qualitatively assess the proposed C-CoT framework. 
In a collision-prone scenario (Fig.~\ref{fig:qual_collision}), the model identifies the critical object and classifies the scene as \textit{Unsafe}. Counterfactual reasoning indicates that maintaining speed or accelerating increases collision risk, while deceleration mitigates it. The model therefore selects deceleration and generates a safe trajectory that avoids the original collision path.

In a safe scenario (Fig.~\ref{fig:qual_safe}), the model infers the scene as \textit{Safe}. Counterfactual analysis shows that acceleration may introduce future risk, whereas excessive deceleration reduces efficiency. Balancing safety and efficiency, the model adopts a maintain-then-decelerate strategy with steering adjustment, producing a trajectory closely aligned with the ground truth.

\section{Conclusion}
This paper presented the C-CoT framework to improve risk reasoning and safety-critical planning in autonomous driving, particularly in complex environments like intersections. By incorporating counterfactual reasoning, the model evaluates alternative meta-actions and their potential consequences within a tree structure, establishing causal links between behaviors and outcomes. Experimental results show that C-CoT outperforms existing models in risk prediction, trajectory planning, and safety. Ablation studies highlight the importance of counterfactual reasoning and meta-action abstraction in improving both safety and interpretability, paving the way for more adaptive and reliable autonomous driving systems.

Future work could extend the framework to lateral maneuvers and design pruning strategies to improve the efficiency of the meta-action evaluation tree.

%\section*{Acknowledgment}

\bibliographystyle{IEEEtran}
\bibliography{ref}

@article{lian2026bap,
  title={BAP-SRL: Bayesian Adaptive Priority Safe Reinforcement Learning for Vehicle Motion Planning at Mixed Traffic Intersections},
  author={Lian, Yuansheng and Zhang, Ke and Guo, Yaming and Li, Shen and Li, Meng},
  journal={arXiv preprint arXiv:2601.21679},
  year={2026}
}

@article{lian2025game,
  title={Game-Theoretic Modeling of Vehicle Unprotected Left Turns Considering Drivers' Bounded Rationality},
  author={Lian, Yuansheng and Zhang, Ke and Li, Meng and Li, Shen},
  journal={arXiv preprint arXiv:2507.03002},
  year={2025}
}

@article{teng2023motion,
  title={Motion planning for autonomous driving: The state of the art and future perspectives},
  author={Teng, Siyu and Hu, Xuemin and Deng, Peng and Li, Bai and others},
  journal={IEEE Transactions on Intelligent Vehicles},
  volume={8},
  number={6},
  pages={3692--3711},
  year={2023},
  publisher={IEEE}
}

@article{tang2022prediction,
  title={Prediction-uncertainty-aware decision-making for autonomous vehicles},
  author={Tang, Xiaolin and Yang, Kai and Wang, Hong and Wu, Jiahang and Qin, Yechen and Yu, Wenhao and Cao, Dongpu},
  journal={IEEE Transactions on Intelligent Vehicles},
  volume={7},
  number={4},
  pages={849--862},
  year={2022},
  publisher={IEEE}
}

@article{hu2026lift,
  title={LIFT: Interpretable truck driving risk prediction with literature-informed fine-tuned LLMs},
  author={Hu, Xiao and Lian, Yuansheng and Li, Meng and Zhang, Ke and Li, Yunxuan and Su, Yuelong},
  journal={Transportation Research Part C: Emerging Technologies},
  volume={185},
  pages={105570},
  year={2026},
  publisher={Elsevier}
}

@inproceedings{zhou2026opendrivevla,
  title={Opendrivevla: Towards end-to-end autonomous driving with large vision language action model},
  author={Zhou, Xingcheng and Han, Xuyuan and Yang, Feng and Ma, Yunpu and Tresp, Volker and Knoll, Alois},
  booktitle={Proceedings of the AAAI Conference on Artificial Intelligence},
  volume={40},
  number={16},
  pages={13782--13790},
  year={2026}
}

@article{cui2025chain,
  title={Chain-of-thought for autonomous driving: A comprehensive survey and future prospects},
  author={Cui, Yixin and Lin, Haotian and Yang, Shuo and Wang, Yixiao and Huang, Yanjun and Chen, Hong},
  journal={arXiv preprint arXiv:2505.20223},
  year={2025}
}

@article{chen2024end,
  title={End-to-end autonomous driving: Challenges and frontiers},
  author={Chen, Li and Wu, Penghao and Chitta, Kashyap and Jaeger, Bernhard and Geiger, Andreas and Li, Hongyang},
  journal={IEEE Transactions on Pattern Analysis and Machine Intelligence},
  volume={46},
  number={12},
  pages={10164--10183},
  year={2024},
  publisher={IEEE}
}

@article{omeiza2021explanations,
  title={Explanations in autonomous driving: A survey},
  author={Omeiza, Daniel and Webb, Helena and Jirotka, Marina and Kunze, Lars},
  journal={IEEE Transactions on Intelligent Transportation Systems},
  volume={23},
  number={8},
  pages={10142--10162},
  year={2021},
  publisher={IEEE}
}

@article{hwang2024emma,
  title={Emma: End-to-end multimodal model for autonomous driving},
  author={Hwang, Jyh-Jing and Xu, Runsheng and Lin, Hubert and Hung, Wei-Chih and Ji, Jingwei and Choi, Kristy and Huang, Di and He, Tong and Covington, Paul and Sapp, Benjamin and others},
  journal={arXiv preprint arXiv:2410.23262},
  year={2024}
}

@inproceedings{sima2024drivelm,
  title={Drivelm: Driving with graph visual question answering},
  author={Sima, Chonghao and Renz, Katrin and Chitta, Kashyap and Chen, Li and Zhang, Hanxue and Xie, Chengen and Bei{\ss}wenger, Jens and Luo, Ping and Geiger, Andreas and Li, Hongyang},
  booktitle={European conference on computer vision},
  pages={256--274},
  year={2024},
  organization={Springer}
}

@inproceedings{nie2024reason2drive,
  title={Reason2drive: Towards interpretable and chain-based reasoning for autonomous driving},
  author={Nie, Ming and Peng, Renyuan and Wang, Chunwei and Cai, Xinyue and Han, Jianhua and Xu, Hang and Zhang, Li},
  booktitle={European Conference on Computer Vision},
  pages={292--308},
  year={2024},
  organization={Springer}
}

@article{long2026vlm,
  title={Vlm-mpc: Model predictive controller augmented vision language model for autonomous driving},
  author={Long, Keke and Shi, Haotian and Liu, Jiaxi and Xiao, Chaowei and Li, Xiaopeng},
  journal={Transportation Research Part C: Emerging Technologies},
  volume={183},
  pages={105487},
  year={2026},
  publisher={Elsevier}
}

@article{xu2024drivegpt4,
  title={Drivegpt4: Interpretable end-to-end autonomous driving via large language model},
  author={Xu, Zhenhua and Zhang, Yujia and Xie, Enze and Zhao, Zhen and Guo, Yong and Wong, Kwan-Yee K and Li, Zhenguo and Zhao, Hengshuang},
  journal={IEEE Robotics and Automation Letters},
  volume={9},
  number={10},
  pages={8186--8193},
  year={2024},
  publisher={IEEE}
}

@inproceedings{shao2024lmdrive,
  title={Lmdrive: Closed-loop end-to-end driving with large language models},
  author={Shao, Hao and Hu, Yuxuan and Wang, Letian and Song, Guanglu and Waslander, Steven L and Liu, Yu and Li, Hongsheng},
  booktitle={Proceedings of the IEEE/CVF conference on computer vision and pattern recognition},
  pages={15120--15130},
  year={2024}
}

@article{bojarski2016end,
  title={End to end learning for self-driving cars},
  author={Bojarski, Mariusz and Del Testa, Davide and Dworakowski, Daniel and Firner, Bernhard and Flepp, Beat and Goyal, Prasoon and Jackel, Lawrence D and Monfort, Mathew and Muller, Urs and Zhang, Jiakai and others},
  journal={arXiv preprint arXiv:1604.07316},
  year={2016}
}

@article{jiang2024senna,
  title={Senna: Bridging large vision-language models and end-to-end autonomous driving},
  author={Jiang, Bo and Chen, Shaoyu and Liao, Bencheng and Zhang, Xingyu and Yin, Wei and Zhang, Qian and Huang, Chang and Liu, Wenyu and Wang, Xinggang},
  journal={arXiv preprint arXiv:2410.22313},
  year={2024}
}

@inproceedings{wang2025omnidrive,
  title={Omnidrive: A holistic vision-language dataset for autonomous driving with counterfactual reasoning},
  author={Wang, Shihao and Yu, Zhiding and Jiang, Xiaohui and Lan, Shiyi and Shi, Min and Chang, Nadine and Kautz, Jan and Li, Ying and Alvarez, Jose M},
  booktitle={Proceedings of the computer vision and pattern recognition conference},
  pages={22442--22452},
  year={2025}
}

@article{peng2025counterfactual,
  title={Counterfactual VLA: Self-Reflective Vision-Language-Action Model with Adaptive Reasoning},
  author={Peng, Zhenghao and Ding, Wenhao and You, Yurong and Chen, Yuxiao and Luo, Wenjie and Tian, Thomas and Cao, Yulong and Sharma, Apoorva and Xu, Danfei and Ivanovic, Boris and others},
  journal={arXiv preprint arXiv:2512.24426},
  year={2025}
}

@inproceedings{wang2020generating,
  title={Generating efficient behaviour with predictive visibility risk for scenarios with occlusions},
  author={Wang, Lingguang and Lopez, Carlos Fetnandez and Stiller, Christoph},
  booktitle={2020 IEEE 23rd International Conference on Intelligent Transportation Systems (ITSC)},
  pages={1--7},
  year={2020},
  organization={IEEE}
}

@article{yu2019occlusion,
  title={Occlusion-aware risk assessment for autonomous driving in urban environments},
  author={Yu, Ming-Yuan and Vasudevan, Ram and Johnson-Roberson, Matthew},
  journal={IEEE Robotics and Automation Letters},
  volume={4},
  number={2},
  pages={2235--2241},
  year={2019},
  publisher={IEEE}
}

@inproceedings{acerbo2021safe,
  title={Safe imitation learning on real-life highway data for human-like autonomous driving},
  author={Acerbo, Flavia Sofia and Alirczaei, Mohsen and Van der Auweraer, Herman and Son, Tong Duy},
  booktitle={2021 IEEE International Intelligent Transportation Systems Conference (ITSC)},
  pages={3903--3908},
  year={2021},
  organization={IEEE}
}

@article{acerbo2022mpc,
  title={MPC-based imitation learning for safe and human-like autonomous driving},
  author={Acerbo, Flavia Sofia and Swevers, Jan and Tuytelaars, Tinne and Son, Tong Duy},
  journal={arXiv preprint arXiv:2206.12348},
  year={2022}
}

@inproceedings{aksjonov2021rule,
  title={Rule-based decision-making system for autonomous vehicles at intersections with mixed traffic environment},
  author={Aksjonov, Andrei and Kyrki, Ville},
  booktitle={2021 IEEE International Intelligent Transportation Systems Conference (ITSC)},
  pages={660--666},
  year={2021},
  organization={IEEE}
}

@article{hu2026long,
  title={Long-and Short-Term Constraint-Driven Safe Reinforcement Learning for Autonomous Driving},
  author={Hu, Xuemin and Chen, Pan and Wen, Yijun and Tang, Bo and Chen, Long},
  journal={IEEE Transactions on Systems, Man, and Cybernetics: Systems},
  year={2026},
  publisher={IEEE}
}

@article{zhou2024vision,
  title={Vision language models in autonomous driving: A survey and outlook},
  author={Zhou, Xingcheng and Liu, Mingyu and Yurtsever, Ekim and Zagar, Bare Luka and Zimmer, Walter and Cao, Hu and Knoll, Alois C},
  journal={IEEE Transactions on Intelligent Vehicles},
  year={2024},
  publisher={IEEE}
}

@inproceedings{wang2024deepaccident,
  title={Deepaccident: A motion and accident prediction benchmark for v2x autonomous driving},
  author={Wang, Tianqi and Kim, Sukmin and Wenxuan, Ji and Xie, Enze and Ge, Chongjian and Chen, Junsong and Li, Zhenguo and Luo, Ping},
  booktitle={Proceedings of the AAAI Conference on Artificial Intelligence},
  volume={38},
  number={6},
  pages={5599--5606},
  year={2024}
}

@inproceedings{liu2024improved,
  title={Improved baselines with visual instruction tuning},
  author={Liu, Haotian and Li, Chunyuan and Li, Yuheng and Lee, Yong Jae},
  booktitle={Proceedings of the IEEE/CVF conference on computer vision and pattern recognition},
  pages={26296--26306},
  year={2024}
}

@article{lee2025efficient,
  title={Efficient llama-3.2-vision by trimming cross-attended visual features},
  author={Lee, Jewon and Song, Ki-Ung and Yang, Seungmin and Lim, Donguk and Kim, Jaeyeon and Shin, Wooksu and Kim, Bo-Kyeong and Lee, Yong Jae and Kim, Tae-Ho},
  journal={arXiv preprint arXiv:2504.00557},
  year={2025}
}

@article{zhu2025internvl3,
  title={Internvl3: Exploring advanced training and test-time recipes for open-source multimodal models},
  author={Zhu, Jinguo and Wang, Weiyun and Chen, Zhe and Liu, Zhaoyang and Ye, Shenglong and Gu, Lixin and Tian, Hao and Duan, Yuchen and Su, Weijie and Shao, Jie and others},
  journal={arXiv preprint arXiv:2504.10479},
  year={2025}
}

@article{yang2025qwen3,
  title={Qwen3 technical report},
  author={Yang, An and Li, Anfeng and Yang, Baosong and Zhang, Beichen and Hui, Binyuan and Zheng, Bo and Yu, Bowen and Gao, Chang and Huang, Chengen and Lv, Chenxu and others},
  journal={arXiv preprint arXiv:2505.09388},
  year={2025}
}

@misc{lu2024deepseekvl,
      title={DeepSeek-VL: Towards Real-World Vision-Language Understanding},
      author={Haoyu Lu and Wen Liu and Bo Zhang and Bingxuan Wang and Kai Dong and Bo Liu and Jingxiang Sun and Tongzheng Ren and Zhuoshu Li and Hao Yang and Yaofeng Sun and Chengqi Deng and Hanwei Xu and Zhenda Xie and Chong Ruan},
      year={2024},
      eprint={2403.05525},
      archivePrefix={arXiv},
      primaryClass={cs.AI}
}

@article{yang2025interactive,
  title={Interactive decision-making integrating graph neural networks and model predictive control for autonomous driving},
  author={Yang, Kai and Li, Shen and Wang, Ming and Tang, Xiaolin},
  journal={IEEE Transactions on Intelligent Transportation Systems},
  volume={26},
  number={5},
  pages={6991--7005},
  year={2025},
  publisher={IEEE}
}

@article{yang2024towards,
  title={Towards safe decision-making for autonomous vehicles at unsignalized intersections},
  author={Yang, Kai and Li, Shen and Chen, Yongli and Cao, Dongpu and Tang, Xiaolin},
  journal={IEEE Transactions on Vehicular Technology},
  volume={74},
  number={3},
  pages={3830--3842},
  year={2024},
  publisher={IEEE}
}

@article{hart2020counterfactual,
  title={Counterfactual policy evaluation for decision-making in autonomous driving},
  author={Hart, Patrick and Knoll, Alois},
  journal={arXiv preprint arXiv:2003.11919},
  year={2020}
}
\end{document}